\documentclass[times, review, 10pt]{elsarticle}

\usepackage{setspace}
\usepackage{amsmath,amssymb,amsfonts}
\usepackage{algorithmic}
\usepackage{graphicx}
\usepackage{textcomp}
\usepackage{booktabs} 
\usepackage{multirow}
\usepackage{caption}
\usepackage{url}
\begin{document}

\begin{frontmatter}

\title{\fontsize{14pt}{16pt}\selectfont HeBA: Heterogeneous Bottleneck Adapters for Robust Vision-Language Models}

\author{Md Jahidul Islam}
\ead{2006123@eee.buet.ac.bd}

\affiliation{organization={Department of Electrical and Electronic Engineering},
            addressline={Bangladesh University of Engineering and Technology}, 
            city={Dhaka},
            country={Bangladesh}}

\begin{abstract}
\fontsize{10pt}{12pt}\selectfont
\doublespacing
Adapting large-scale Vision-Language Models (VLMs) like CLIP to downstream tasks often suffers from a "one-size-fits-all" architectural approach, where visual and textual tokens are processed uniformly by wide, generic adapters. We argue that this homogeneity ignores the distinct structural nature of the modalities—spatial locality in images versus semantic density in text. To address this, we propose \textbf{HeBA} (Heterogeneous Bottleneck Adapter), a unified architectural framework that introduces modality-specific structural inductive biases. HeBA departs from conventional designs through three key architectural innovations: (1) \textbf{Heterogeneity:} It processes visual tokens via 2D depthwise-separable convolutions to preserve spatial correlations, while distinctively processing text tokens via dense linear projections to capture semantic relationships; (2) \textbf{Bottleneck Regularization:} Unlike standard expanding adapters, HeBA employs a compression bottleneck ($D \rightarrow D/4$) that explicitly forces the model to learn compact, robust features and acts as a structural regularizer; and (3) \textbf{Active Gradient Initialization:} We challenge the restrictive zero-initialization paradigm, utilizing a Kaiming initialization strategy that ensures sufficient initial gradient flow to accelerate convergence without compromising the frozen backbone's pre-trained knowledge. Extensive experiments demonstrate that HeBA's architecturally specialized design achieves superior stability and accuracy, establishing a new state-of-the-art on 11 few-shot benchmarks. Code is available at \url{https://github.com/Jahid12012021/VLM-HeBA}.
\end{abstract}


\end{frontmatter}

\section{Introduction}
Vision-Language Models (VLMs), exemplified by CLIP~\cite{radford2021learning}, ALIGN~\cite{jia2021scaling}, and Florence~\cite{yuan2021florence}, have fundamentally reshaped the landscape of computer vision. By pre-training on billion-scale datasets of noisy image-text pairs via contrastive learning, these models align visual and semantic representations in a unified embedding space. This alignment grants them unprecedented zero-shot generalization capabilities, allowing them to recognize arbitrary concepts without task-specific training. However, despite their robustness, deploying VLMs in downstream applications often requires adaptation to specific domains (e.g., satellite imagery, medical scans) where the pre-training distribution differs significantly from the target distribution~\cite{zhang2022tip, radford2021learning}.

Adapting these large-scale models with limited data—a setting known as \textit{few-shot learning}—presents a formidable ``Stability-Plasticity'' dilemma. Naive fine-tuning of the entire backbone is computationally prohibitive and prone to \textit{catastrophic forgetting}, where the model overfits to the few training examples (Base classes) and aggressively degrades on unseen categories (Novel classes)~\cite{zhou2022learning}. Consequently, research has pivoted toward Parameter-Efficient Fine-Tuning (PEFT), which freezes the backbone and injects lightweight learnable modules.

\begin{figure}[t]
  \centering
  \includegraphics[width=\linewidth]{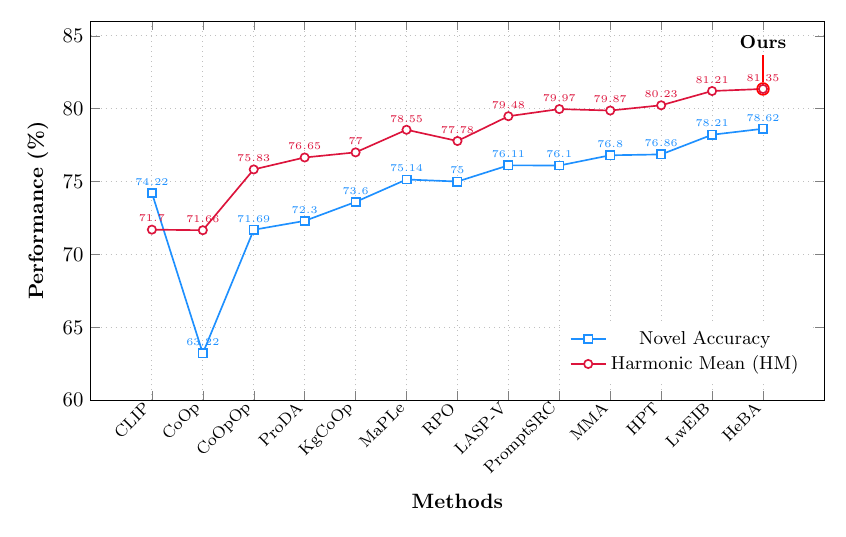}
  \caption{\doublespacing
  \textbf{Chronological Base-to-Novel Generalization.} Novel Accuracy (blue) and Harmonic Mean (red) across 11 datasets. \textbf{HeBA (Ours)} sets a new state-of-the-art with \textbf{78.62\% Novel Accuracy} and \textbf{81.35\% HM}.}
  \label{fig:chronological_performance}
\end{figure}

Existing PEFT approaches generally fall into two categories: \textit{Prompt Learning} and \textit{Adapter Tuning}. Prompt learning methods, such as CoOp~\cite{zhou2022learning} and MaPLe~\cite{khattak2023maple}, optimize learnable tokens in the text or multimodal encoders. While effective for semantic alignment, these methods often struggle to capture fine-grained spatial details, as they operate primarily on global token representations~\cite{khattak2023self, zhu2023prompt}. Conversely, Adapter-based methods, such as CLIP-Adapter~\cite{gao2023clip} and Tip-Adapter~\cite{zhang2022tip}, insert Multi-Layer Perceptrons (MLPs) into the image encoder. However, a critical limitation persists: most current adapters suffer from \textit{architectural homogeneity}. They treat visual tokens (which possess intrinsic 2D spatial correlations) and textual tokens (which are dense semantic sequences) as uniform 1D vectors~\cite{yang2025learning}. This ``spatial amnesia'' often discards critical structural cues—such as textures in satellite imagery or shapes in fine-grained classification—limiting adaptation performance~\cite{gao2023clip, zhang2022tip}.

Recent state-of-the-art methods like LwEIB~\cite{yang2025learning} attempt to reintroduce spatial inductive biases by incorporating depthwise convolutions. However, their architectural design relies on "Inverse Bottlenecks" that expand the internal feature dimension to four times the input width ($4\times$), significantly increasing parameter count and overfitting risks in data-scarce regimes. While LwEIB employs a stochastic ``slow-fast'' optimization schedule to manage this volatility, applying such dynamic scaling to an unconstrained, high-capacity architecture creates a fragile optimization landscape where convergence becomes highly sensitive to hyperparameter tuning. We argue that dynamic optimization strategies should not serve as remedial tools for architectural instability. Instead, they function best when paired with structural regularization—specifically, compressive bottlenecks—shifting the role of the optimization schedule from mere stabilization to maximizing feature adaptation efficiency.

In this work, we introduce \textbf{HeBA} (Heterogeneous Bottleneck Adapter), a unified framework that resolves these issues by encoding domain-specific priors directly into the architecture. Unlike prior works that rely on homogeneous layers or parameter-heavy expansions, HeBA distinguishes itself through three key synergistic contributions:

\begin{enumerate}
    \item \textbf{Heterogeneous Inductive Biases:} We argue that vision and language require distinct processing pipelines. HeBA employs a bifurcated architecture: a \textit{Visual Stream} utilizing 2D depthwise-separable convolutional bottlenecks to explicitly model spatial locality~\cite{chollet2017xception, he2016deep}, and a \textit{Textual Stream} utilizing dense linear bottlenecks to preserve global semantic integrity~\cite{vaswani2017attention}. This heterogeneity ensures that structural correlations are preserved for images while semantic density is maintained for text.

    \item \textbf{Structural Regularization via Bottlenecks:} We demonstrate that the architecture itself can act as a regularizer. HeBA replaces the standard expanding adapter design with a compressive \textit{Bottleneck Structure} ($D \rightarrow D/4$)~\cite{hu2021lora}. This constraint restricts the model's capacity to overfit, forcing it to learn a low-rank, compact representation of the domain shift, physically filtering out task-irrelevant noise without the need for complex external regularizers.

    \item \textbf{Active Gradient Initialization Paradigm:} Challenging the prevailing consensus in PEFT methods like MaPLe~\cite{khattak2023maple} and Tip-Adapter~\cite{zhang2022tip}, which rely on zero-initialization to strictly preserve identity mappings, we introduce an \textbf{Active Kaiming Initialization} strategy~\cite{he2015delving}. While zero-initialization can lead to vanishing gradients in the adapter layers during early training stages, our strategy ensures sufficient initial gradient magnitude to rapidly adapt to the downstream distribution. Coupled with dynamic scaling and \textit{Label Smoothing}~\cite{szegedy2016rethinking} to stabilize this active learning phase, this approach achieves superior convergence and sets a new state-of-the-art Harmonic Mean of \textbf{81.35\%} across 11 benchmarks.
\end{enumerate}

\section{Related Work}

\subsection{Vision-Language Models and Adaptation}
The advent of Vision-Language Models (VLMs) like CLIP~\cite{radford2021learning} and ALIGN~\cite{jia2021scaling} has shifted the paradigm from training task-specific models to adapting general-purpose foundations. While full fine-tuning~\cite{wortsman2022robust} can update all parameters, it often destroys the pre-trained feature space, leading to poor Out-of-Distribution (OOD) generalization. Consequently, research has pivoted to Parameter-Efficient Fine-Tuning (PEFT), aiming to adapt models with minimal parameter updates while preserving zero-shot robustness.

\subsection{Prompt Learning}
Inspired by NLP, prompt learning optimizes the input text tokens while keeping the backbone frozen. \textbf{CoOp}~\cite{zhou2022learning} replaced manual templates with learnable continuous vectors. While effective for Base classes, it suffered from overfitting on Novel classes. \textbf{CoCoOp}~\cite{zhou2022conditional} addressed this by conditioning prompts on image instances via a meta-network. \textbf{ProDA}~\cite{lu2022prompt} further improved generalization by learning the distribution of prompts rather than a single vector.

Recent works focus on semantic alignment and regularization. \textbf{KgCoOp}~\cite{yao2023visual} minimizes the discrepancy between learnable and handcrafted prompts to retain general knowledge. \textbf{MaPLe}~\cite{khattak2023maple} introduced multi-modal prompting, injecting learnable tokens into both vision and language branches to ensure deep alignment. Other approaches focus on regularization constraints: \textbf{PromptSRC}~\cite{khattak2023self} uses self-regularization to prevent forgetting, \textbf{RPO}~\cite{lee2023read} optimizes special read-only tokens with masking strategies, and \textbf{LASP-V}~\cite{bulat2023lasp} employs language-aware soft prompting to regularize the text encoder using distinct visual-language losses.

\subsection{Adapter-Based and Hybrid Approaches}
Adapters insert lightweight residual modules into the frozen backbone. \textbf{CLIP-Adapter}~\cite{gao2023clip} appends a bottleneck MLP to the encoders to refine features. \textbf{Tip-Adapter}~\cite{zhang2022tip} constructs a key-value cache from few-shot examples for training-free adaptation.

More recent methods leverage auxiliary knowledge or cross-modal interactions. \textbf{HPT}~\cite{wang2024learning} utilizes Large Language Models (LLMs) to generate hierarchical descriptions to structure the semantic space. \textbf{MMA} (Multi-Modal Adapter)~\cite{yang2024mma} proposes a dual-pathway adapter that bridges visual and textual features through cross-modal attention.

The direct predecessor to our work, \textbf{LwEIB}~\cite{yang2025learning}, introduced depthwise convolutions but relied on an ``Inverse Bottleneck'' design that expands the internal feature dimension ($4\times$). This parameter-heavy approach necessitates heuristic optimization schedules to prevent representational collapse. \textbf{HeBA} distinguishes itself by inverting this architectural logic: we employ \textbf{Heterogeneous Bottleneck Adapters} that compress features ($D \rightarrow D/4$). This architecture serves as an intrinsic structural regularizer, ensuring representational stability by design. Consequently, it permits active gradient initialization and dynamic optimization without the severe risk of divergence associated with over-parameterized modules.

\begin{figure}[t]
  \centering
  \includegraphics[width=\textwidth]{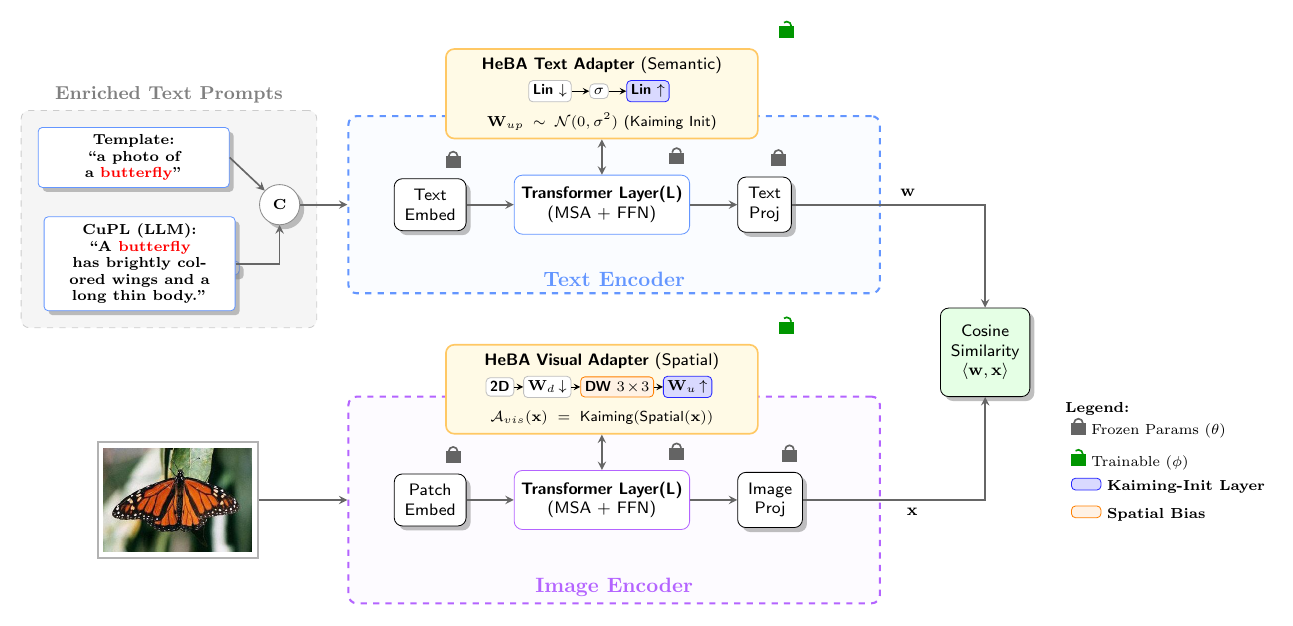}
  \caption{\doublespacing
  \textbf{Overview of the HeBA framework.} We keep the pre-trained CLIP~\cite{radford2021learning} backbone frozen (indicated by \includegraphics[height=0.8em]{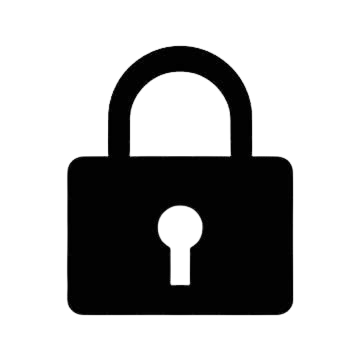}) and inject lightweight, modality-specific adapters. 
  \textbf{Left:} Enriched text prompts combine standard handcrafted templates with fine-grained LLM descriptions (CuPL) to enhance semantic representation. 
  \textbf{Top:} The Text Adapter employs a \textbf{Bottleneck} linear architecture to preserve semantic integrity while compressing dimensions. 
  \textbf{Bottom:} The Visual Adapter explicitly captures spatial inductive biases using $3 \times 3$ depthwise convolutions (DW-Conv). 
  \textbf{Key Innovation:} Unlike prior methods, the up-projection layers utilize \textbf{Active Kaiming Initialization} to provide immediate gradient flow, driving rapid feature adaptation from the first iteration and mitigating zero-gradient stagnation.}
  \label{fig:heba_arch}
\end{figure}
\subsection{Inductive Biases in Few-Shot Learning}
Inductive biases are critical for sample efficiency. While CNNs enforce locality~\cite{he2016deep} and Transformers enforce global attention~\cite{dosovitskiy2020image}, few-shot adapters often lack explicit structural constraints. \textbf{HeBA} explicitly decouples these biases: we enforce \textbf{2D Spatial Locality} for the visual stream via depthwise-separable convolutions and \textbf{Semantic Globalism} for the text stream via linear projections. By aligning the adapter architecture with the intrinsic structure of the data, HeBA achieves superior efficiency compared to modality-agnostic or purely prompt-based approaches.

\section{Methodology}

We introduce \textbf{HeBA} (Heterogeneous Bottleneck Adapter), a unified architectural framework designed to robustly adapt the frozen CLIP backbone~\cite{radford2021learning} to downstream tasks. HeBA departs from the expansive design of prior spatial adapters like LwEIB~\cite{yang2025learning} by enforcing strict dimension compression coupled with modality-specific processing.

\subsection{Heterogeneous Bottleneck Architecture}
Let the input feature sequence at layer $l$ be denoted as $\mathbf{x}_l \in \mathbb{R}^{N \times D}$, where $N$ is the sequence length and $D$ is the embedding dimension. The adapted output $\mathbf{x}_{l+1}$ is computed via a residual connection:
\begin{equation}
    \mathbf{x}_{l+1} = \text{LayerNorm}\left( \mathbf{x}_l + s \cdot \mathcal{F}_{HeBA}(\mathbf{x}_l) \right)
\end{equation}
where $\text{LayerNorm}$ denotes layer normalization~\cite{ba2016layer} and $s$ is a dynamic scaling factor. Unlike LwEIB, which expands the internal dimension to $4D$, HeBA employs a compressive bottleneck that projects features down to $D' = D/r$ (with reduction ratio $r=4$). This compression acts as a structural regularizer, forcing the adapter to isolate and learn a low-rank representation of the domain shift.
\begin{figure*}[htbp]
  \centering
  \includegraphics[width=\linewidth]{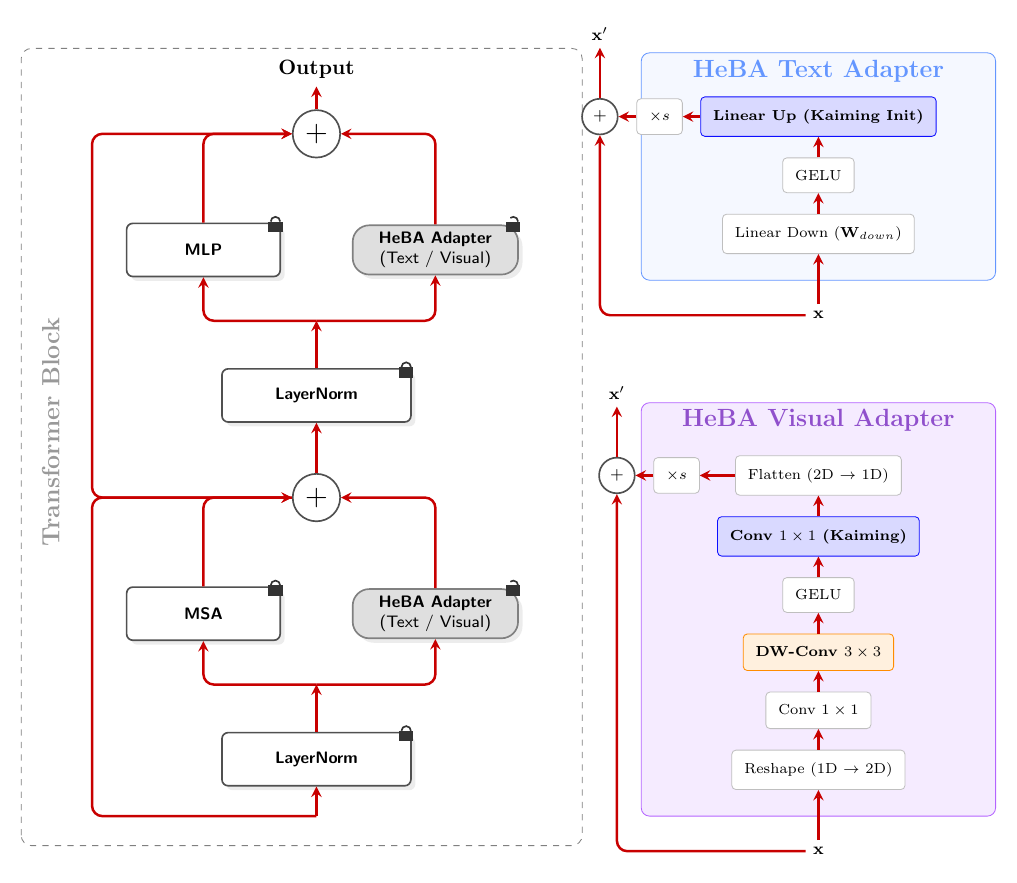}
  \caption{\doublespacing
  \textbf{Model-level Inductive Bias Integration in HeBA.} 
  \textbf{Left:} Parallel adapters are inserted into the frozen Transformer block (MSA and MLP) to learn residual corrections. 
  \textbf{Right:} The \textbf{Text Adapter} utilizes a linear bottleneck ($D \rightarrow D/4$) to preserve semantics, while the \textbf{Visual Adapter} employs $3\times3$ Depthwise Convolutions to enforce spatial locality. Crucially, up-projections use \textbf{Kaiming Initialization} to actively stimulate learning, modulated by a dynamic scaling factor $s$.}
  \label{fig:heba_structure}
\end{figure*}
\subsubsection{Visual Stream: Spatial-Aware Convolution}
Visual tokens in CLIP possess intrinsic 2D spatial correlations that are lost when treated as flat sequences. To preserve this geometry, we employ a heterogeneous design for the visual branch. We first reshape the input tokens into a 2D grid $\mathbf{X}_{2D} \in \mathbb{R}^{B \times D \times \sqrt{N} \times \sqrt{N}}$. The visual adapter function $\mathcal{F}_{vis}$ is defined as a sequence of specialized convolutions:
\begin{equation}
    \mathbf{Z}_{down} = \text{Conv}_{1\times1}(\mathbf{X}_{2D}) \quad \in \mathbb{R}^{B \times \frac{D}{r} \times \sqrt{N} \times \sqrt{N}}
\end{equation}
\begin{equation}
    \mathbf{Z}_{mid} = \text{DW-Conv}_{3\times3}(\mathbf{Z}_{down})
\end{equation}
\begin{equation}
    \mathcal{F}_{vis}(\mathbf{x}) = \text{Flatten}\left( \text{Conv}_{1\times1}\left( \sigma(\mathbf{Z}_{mid}) \right) \right)
\end{equation}
Here, $\text{Conv}_{1\times1}$ performs channel-wise compression, and $\text{DW-Conv}_{3\times3}$ aggregates local spatial context. The activation function $\sigma(\cdot)$ is the Gaussian Error Linear Unit (GELU)~\cite{hendrycks2016gaussian}, chosen for its smooth probabilistic properties. This design explicitly models spatial locality (e.g., textures, shapes) critical for visual recognition.

\subsubsection{Textual Stream: Semantic-Preserving Projection}
For the textual stream, spatial locality is irrelevant. Therefore, HeBA switches to a dense linear topology to preserve global semantic integrity. The textual adapter function $\mathcal{F}_{text}$ operates directly on the token sequence:
\begin{equation}
    \mathcal{F}_{text}(\mathbf{x}) = \mathbf{W}_{up} \cdot \sigma\left( \mathbf{W}_{down} \cdot \mathbf{x} \right)
\end{equation}
where $\mathbf{W}_{down} \in \mathbb{R}^{D \times \frac{D}{r}}$ and $\mathbf{W}_{up} \in \mathbb{R}^{\frac{D}{r} \times D}$ are linear projection matrices. By avoiding spatial convolutions for text, HeBA respects the distinct structural nature of linguistic data.

\subsection{Active Gradient Initialization Paradigm}
A critical theoretical divergence of HeBA lies in its initialization strategy. Prevailing PEFT methods, such as Tip-Adapter~\cite{zhang2022tip} and MaPLe~\cite{khattak2023maple}, explicitly initialize their adaptation modules with zeros (setting $\mathbf{W}_{up} = 0$). The motivation behind this is to preserve a strict identity mapping at the onset of training, theoretically ensuring that the pre-trained knowledge of the original CLIP model is perfectly retained.

However, we argue that this zero-initialization induces a prolonged state of vanishing gradients within the newly introduced adapter subspace, artificially delaying the model's ability to adapt to severe distribution shifts. To overcome this, HeBA introduces an \textbf{Active Kaiming Initialization} strategy~\cite{he2015delving}:
\begin{equation}
    \mathbf{W}_{up} \sim \mathcal{N}(0, \frac{2}{n_{in}}), \quad \mathbf{b}_{up} = 0
\end{equation}
By initializing the weights with a He Normal distribution, we ensure an immediate and robust gradient flow from the very first iteration ($t=0$). Because the primary CLIP backbone remains strictly frozen, the core pre-trained knowledge is intrinsically safe from catastrophic forgetting. Thus, this active initialization provides the necessary momentum for the adapters to rapidly map out domain-specific residuals, preventing the optimizer from stagnating in the pre-trained model's local minimum.

\subsection{Optimization and Regularization}
To theoretically balance the active adaptation initiated by the Kaiming initialization and prevent potential divergence, we employ two complementary regularization mechanisms:

\textbf{1. Dynamic Slow-Fast Schedule:} To navigate the complex optimization landscape and escape local saddle points, we employ a stochastic scaling mechanism. The adapter's output scale factor $s$ is randomly amplified with probability $p$:
\begin{equation}
    s_{train} = s \cdot (1 + \mathbb{I}_{u < p} \cdot \alpha)
\end{equation}
where $\alpha$ is the scaling factor and $u \sim U(0,1)$. This dynamic scaling acts as a stabilizing force, complementing the active initialization by carefully modulating the magnitude of the adapter's influence during training.

\textbf{2. Label Smoothing:} To prevent the model from generating overconfident predictions on the limited few-shot examples, we replace the standard Cross-Entropy loss with Label Smoothing Cross-Entropy (LSCE)~\cite{szegedy2016rethinking}:
\begin{equation}
    \mathcal{L}_{LSCE} = (1 - \epsilon) \mathcal{L}_{CE} + \epsilon \frac{1}{K} \sum_{k=1}^{K} - \log(p_k)
\end{equation}
where $\epsilon=0.1$ is the smoothing parameter. This theoretically penalizes peaky probability distributions, significantly enhancing generalization to unseen Novel classes.

\section{Experiments}
\subsection{Experimental Setup}

\noindent\textbf{Generalization from Base-to-Novel Classes.}
Following the established protocol in CoOp~\cite{zhou2022learning}, we evaluate HeBA on 11 diverse image classification datasets covering general objects (ImageNet~\cite{deng2009imagenet}, Caltech101~\cite{feifei2004learning}), fine-grained categories (OxfordPets~\cite{parkhi2012cats}, StanfordCars~\cite{krause20133d}, Flowers102~\cite{nilsback2008automated}, Food101~\cite{bossard2014food}, FGVCAircraft~\cite{maji2013fine}), scenes (SUN397~\cite{xiao2010sun}), textures (DTD~\cite{cimpoi2014describing}), satellite imagery (EuroSAT~\cite{helber2019eurosat}), and actions (UCF101~\cite{soomro2012ucf101}).
We split the classes into two disjoint groups: Base (seen) and Novel (unseen). The model is trained on Base classes using 16 shots per category and evaluated on both Base and Novel classes. We report the accuracy for both groups and their Harmonic Mean (HM) to measure the trade-off between adaptation and generalization.

\noindent\textbf{Cross-Dataset Evaluation.}
To assess transferability, we train our model on ImageNet (16 shots per class) using all 1,000 classes. We then evaluate the trained model directly on the remaining 10 datasets without any further fine-tuning, following the protocol in CoCoOp~\cite{zhou2022conditional}.

\noindent\textbf{Domain Generalization.}
To evaluate robustness against distribution shifts, we use the model trained on ImageNet and test it on four out-of-distribution variants: ImageNetV2~\cite{recht2019imagenet}, ImageNet-Sketch~\cite{wang2019learning}, ImageNet-A~\cite{hendrycks2021natural}, and ImageNet-R~\cite{hendrycks2021many}.

\subsection{Implementation Details}
We implement HeBA using the ViT-B/16 CLIP backbone~\cite{radford2021learning}. The image encoder and text encoder are kept frozen, and only the HeBA adapter parameters are updated.

\textbf{Architecture:} We utilize a heterogeneous design to respect modality-specific structures. The visual adapter employs depthwise-separable convolutions with a kernel size of $3 \times 3$ to explicitly capture local spatial geometry~\cite{chollet2017xception}, while the text adapter utilizes linear projections to maintain semantic integrity. Unlike prior expansion-based methods, HeBA enforces a \textbf{bottleneck reduction ratio} of $r=4$ (compressing dimension $D \rightarrow D/4$) to act as a structural regularizer against overfitting.

\textbf{Optimization:} We employ a \textbf{Kaiming Initialization} strategy~\cite{he2015delving} for the up-projection weights to enable an active initial gradient flow, effectively avoiding the delayed convergence often associated with zero-initialization. The model is trained using the AdamW optimizer with a learning rate of $1 \times 10^{-3}$, utilizing a stochastic ``slow-fast'' schedule~\cite{yang2025learning} to modulate adapter scaling during training. The objective function is regularized via \textbf{Label Smoothing Cross-Entropy} with $\epsilon=0.1$~\cite{szegedy2016rethinking}.

\textbf{Prompts:} We utilize the standard template ``a photo of a \{class\}'' enriched with LLM-generated descriptions from CuPL~\cite{pratt2023what}. Following established protocols, we utilize multiple descriptions per category to robustly represent the semantic space. \\
\noindent\textbf{Training Configuration.}
We use SGD optimizer and a cosine annealing learning rate scheduler followed by the LwEIB~\cite{yang2025learning}. All experiments are conducted on a single ``NVIDIA Tesla P100 GPU'' (via Kaggle Kernels).
\begin{itemize}
    \item \textbf{Base-to-Novel Generalization:} We train for 30 epochs with a batch size of 16. To ensure stability, we use a conservative learning rate of $7.5 \times 10^{-3}$. The adapter scaling factor is set to $\alpha_{base} = 0.025$ with a multiplier $s=2.25$. We employ a negative sampling ratio of 5 and a slow-fast ratio of 0.8~\cite{yang2025learning}. Crucially, during inference on Novel classes, we adjust the adapter scale to $\alpha_{novel} = 0.010$ to prevent overfitting to the base class statistics, while keeping $\alpha_{base} = 0.025$.
    \item \textbf{Cross-Dataset \& Domain Generalization:} Following MaPLe~\cite{khattak2023maple}, optimization is performed using SGD with a momentum of 0.9 and a weight decay of 0.0005 and we train for 10 epochs with a batch size of 64 and a learning rate of $6.5 \times 10^{-3}$. The scaling factor is set to $\alpha_{base} = 0.05$ and $\alpha_{novel} = 0.025$ with a multiplier $s=10.0$.
\end{itemize}
All results are reported as the average over three independent runs with different random seeds (1, 2, 3).

\section{Results}

\subsection{Generalization from Base-to-Novel Classes}
We compare HeBA against state-of-the-art methods on the Base-to-Novel generalization setting. The results are summarized in Table \ref{tab:sota_comparison}.

\textbf{Analysis.} HeBA achieves a new state-of-the-art harmonic mean (HM) of \textbf{81.35\%}, surpassing the strong baseline LwEIB (81.21\%)~\cite{yang2025learning} and MMA (79.87\%)~\cite{yang2024mma}. A key highlight is HeBA's superior generalization to novel classes, achieving \textbf{78.62\%} accuracy compared to LwEIB's 78.21\%. This demonstrates that our compressive structural bottleneck ($D \rightarrow D/4$) effectively mitigates the overfitting susceptibility inherent to expanding adapters and prompt learning methods like CoOp (63.22\% Novel)~\cite{zhou2022learning}.

HeBA exhibits notable proficiency in structure-sensitive and domain-shifted datasets. On \textbf{DTD} (textures)~\cite{cimpoi2014describing}, HeBA improves novel accuracy by \textbf{+2.37\%} over LwEIB (70.20\% vs 67.83\%). Similarly, on \textbf{EuroSAT} (satellite imagery)~\cite{helber2019eurosat}, HeBA achieves a harmonic mean of \textbf{88.16\%}, outperforming LwEIB (86.86\%). This empirical evidence validates our theoretical assertion that explicit 2D spatial modeling via depthwise convolutions is paramount for recognizing fine-grained geometric patterns in non-object-centric domains.

\begin{table*}[htbp]
\centering
\caption{\doublespacing
Comparison with state-of-the-art methods on Base-to-Novel Generalization (Part 1/3). 
HeBA achieves the highest average Harmonic Mean (HM).}
\label{tab:sota_comparison}
\footnotesize
\setlength{\tabcolsep}{3pt} 
\begin{tabular}{l c ccc ccc ccc ccc}
\toprule
\multirow{2}{*}{\textbf{Method}} & \multirow{2}{*}{\textbf{Year}} & \multicolumn{3}{c}{\textbf{Average (11)}} & \multicolumn{3}{c}{\textbf{ImageNet}} & \multicolumn{3}{c}{\textbf{Caltech101}} & \multicolumn{3}{c}{\textbf{Oxford Pets}} \\
 & & Base & Novel & HM & Base & Novel & HM & Base & Novel & HM & Base & Novel & HM \\
\midrule
CLIP & 2021 & 69.34 & 74.22 & 71.70 & 72.43 & 68.14 & 70.22 & 96.84 & 94.00 & 95.40 & 91.17 & 97.26 & 94.12 \\
CoOp & 2022 & 82.69 & 63.22 & 71.66 & 76.47 & 67.88 & 71.92 & 98.00 & 89.81 & 93.73 & 93.67 & 95.29 & 94.47 \\
CoOpOp & 2022 & 80.47 & 71.69 & 75.83 & 75.98 & 70.43 & 73.10 & 97.96 & 93.81 & 95.84 & 95.20 & 97.69 & 96.43 \\
ProDA & 2022 & 81.56 & 72.30 & 76.65 & 75.40 & 70.23 & 72.72 & 98.27 & 93.23 & 95.68 & 95.43 & 97.83 & 96.62 \\
KgCoOp & 2023 & 80.73 & 73.60 & 77.00 & 75.83 & 69.96 & 72.78 & 97.72 & 94.39 & 96.03 & 94.65 & 97.76 & 96.18 \\
MaPLe & 2023 & 82.28 & 75.14 & 78.55 & 76.66 & 70.54 & 73.47 & 97.74 & 94.36 & 96.02 & 95.43 & 97.76 & 96.58 \\
LASP-V & 2023 & 83.18 & 76.11 & 79.48 & 76.25 & 71.17 & 73.62 & 98.17 & 94.33 & 96.21 & 95.73 & 97.87 & 96.79 \\
RPO & 2023 & 81.13 & 75.00 & 77.78 & 76.60 & 71.57 & 74.00 & 97.97 & 94.37 & 96.03 & 94.63 & 97.50 & 96.05 \\
P-SRC & 2023 & 84.26 & 76.10 & 79.97 & 77.60 & 70.73 & 74.01 & 98.10 & 94.03 & 96.02 & 95.33 & 97.30 & 96.30 \\
HPT & 2024 & 84.32 & 76.86 & 80.23 & \textbf{77.95} & 70.74 & 74.17 & 98.37 & 94.98 & 96.65 & 95.78 & 97.65 & 96.71 \\
MMA & 2024 & 83.20 & 76.80 & 79.87 & 77.31 & 71.00 & 74.02 & 98.40 & 94.00 & 96.15 & 95.40 & \textbf{98.07} & \textbf{96.72} \\
LwEIB & 2025 & \textbf{84.45} & 78.21 & 81.21 & 76.64 & 71.64 & 74.06 & \textbf{98.47} & 95.47 & \textbf{96.95} & 95.70 & 97.40 & 96.54 \\
\textbf{HeBA} & \textbf{Ours} & 84.29 & \textbf{78.62} & \textbf{81.35} & 77.53 & \textbf{71.53} & \textbf{74.41} & 98.41 & \textbf{95.48} & 96.92 & \textbf{95.71} & 97.00 & 96.35 \\
\bottomrule
\end{tabular}

\vspace{10pt}

\begin{tabular}{l c ccc ccc ccc ccc}
\toprule
\multirow{2}{*}{\textbf{Method}} & \multirow{2}{*}{\textbf{Year}} & \multicolumn{3}{c}{\textbf{Stan. Cars}} & \multicolumn{3}{c}{\textbf{Flowers}} & \multicolumn{3}{c}{\textbf{Food101}} & \multicolumn{3}{c}{\textbf{FGVC Aircraft}} \\
 & & Base & Novel & HM & Base & Novel & HM & Base & Novel & HM & Base & Novel & HM \\
\midrule
CLIP & 2021 & 63.37 & 74.89 & 68.65 & 72.08 & 77.80 & 74.83 & 90.10 & 91.22 & 90.66 & 27.19 & 36.29 & 31.09 \\
CoOp & 2022 & 78.12 & 60.40 & 68.13 & 97.60 & 59.67 & 74.06 & 88.33 & 82.26 & 85.19 & 40.44 & 22.30 & 28.75 \\
CoOpOp & 2022 & 70.49 & 73.59 & 72.01 & 94.87 & 71.75 & 81.71 & 90.70 & 91.29 & 90.99 & 33.41 & 23.71 & 27.74 \\
ProDA & 2022 & 74.70 & 71.20 & 72.91 & 97.70 & 68.68 & 80.66 & 90.30 & 88.57 & 89.43 & 36.90 & 34.13 & 35.46 \\
KgCoOp & 2023 & 71.76 & 75.04 & 73.36 & 95.00 & 74.73 & 83.65 & 90.50 & 91.70 & 91.09 & 36.21 & 33.55 & 34.83 \\
MaPLe & 2023 & 72.94 & 74.00 & 73.47 & 95.92 & 72.46 & 82.56 & 90.71 & \textbf{92.05} & 91.38 & 37.44 & 35.61 & 36.50 \\
LASP-V & 2023 & 75.23 & 71.77 & 73.46 & 97.17 & 73.53 & 83.71 & \textbf{91.20} & 91.90 & \textbf{91.54} & 38.05 & 33.20 & 35.46 \\
RPO & 2023 & 73.87 & 75.53 & 74.69 & 94.13 & 76.67 & 84.50 & 90.33 & 90.83 & 90.58 & 37.33 & 34.20 & 35.70 \\
P-SRC & 2023 & 78.27 & 74.97 & 76.58 & 98.07 & 76.50 & 85.95 & 90.67 & 91.53 & 91.10 & 42.73 & 37.87 & 40.15 \\
HPT & 2024 & 76.95 & 74.23 & 75.57 & \textbf{98.17} & \textbf{78.37} & \textbf{87.16} & 90.46 & 91.57 & 91.01 & 42.68 & 38.13 & 40.28 \\
MMA & 2024 & 78.50 & 73.10 & 75.70 & 97.77 & 75.93 & 85.48 & 90.13 & 90.71 & 91.30 & 36.33 & 40.57 & 38.33 \\
LwEIB & 2025 & \textbf{80.07} & 74.01 & 76.92 & 97.53 & 77.50 & 86.37 & 90.63 & 91.73 & 91.18 & \textbf{45.11} & \textbf{42.60} & \textbf{43.82} \\
\textbf{HeBA} & \textbf{Ours} & 78.80 & \textbf{75.94} & \textbf{77.34} & 97.37 & \textbf{78.37} & 86.84 & 90.55 & 91.66 & 91.10 & 42.38 & 40.71 & 41.53 \\
\bottomrule
\end{tabular}

\vspace{10pt}

\begin{tabular}{l c ccc ccc ccc ccc}
\toprule
\multirow{2}{*}{\textbf{Method}} & \multirow{2}{*}{\textbf{Year}} & \multicolumn{3}{c}{\textbf{SUN397}} & \multicolumn{3}{c}{\textbf{DTD}} & \multicolumn{3}{c}{\textbf{EuroSAT}} & \multicolumn{3}{c}{\textbf{UCF101}} \\
 & & Base & Novel & HM & Base & Novel & HM & Base & Novel & HM & Base & Novel & HM \\
\midrule
CLIP & 2021 & 69.36 & 75.35 & 72.23 & 53.24 & 59.90 & 56.37 & 56.48 & 64.05 & 60.03 & 70.53 & 77.50 & 73.85 \\
CoOp & 2022 & 80.60 & 65.89 & 72.51 & 79.44 & 41.18 & 54.24 & 92.19 & 54.74 & 68.69 & 84.69 & 56.05 & 67.46 \\
CoOpOp & 2022 & 79.74 & 76.86 & 78.27 & 77.01 & 56.00 & 64.85 & 87.49 & 60.04 & 71.21 & 82.33 & 73.45 & 77.64 \\
ProDA & 2022 & 78.67 & 76.93 & 77.79 & 80.67 & 56.48 & 66.44 & 83.90 & 66.00 & 73.88 & 85.23 & 71.97 & 78.04 \\
KgCoOp & 2023 & 80.29 & 76.53 & 78.36 & 77.55 & 54.99 & 64.35 & 85.64 & 64.34 & 73.48 & 82.89 & 76.67 & 79.65 \\
MaPLe & 2023 & 80.82 & 78.70 & 79.75 & 80.36 & 59.18 & 68.16 & 94.07 & 73.23 & 82.35 & 78.66 & 80.77 & 83.00 \\
LASP-V & 2023 & 80.70 & 79.30 & 80.00 & 81.10 & 62.57 & 70.64 & 95.00 & \textbf{83.37} & \textbf{88.86} & 85.53 & 78.20 & 81.70 \\
RPO & 2023 & 80.60 & 77.80 & 79.18 & 76.70 & 62.13 & 68.61 & 86.63 & 68.97 & 76.79 & 83.67 & 75.43 & 79.34 \\
P-SRC & 2023 & \textbf{82.67} & 78.47 & 80.52 & 83.37 & 62.97 & 71.75 & 92.90 & 73.90 & 82.32 & \textbf{87.10} & 78.80 & 82.74 \\
HPT & 2024 & 82.57 & 79.26 & \textbf{80.88} & \textbf{83.84} & 63.33 & 72.16 & 94.24 & 77.12 & 84.82 & 86.52 & 80.06 & 83.16 \\
MMA & 2024 & 82.27 & 78.57 & 80.38 & 83.20 & 65.63 & 73.38 & 85.46 & 82.34 & 83.87 & 86.23 & 80.03 & 82.20 \\
LwEIB & 2025 & 81.10 & \textbf{79.80} & 80.44 & 82.87 & 67.83 & 74.60 & 95.00 & 80.01 & 86.86 & 85.73 & 82.37 & 84.02 \\
\textbf{HeBA} & \textbf{Ours} & 81.90 & 79.30 & 80.58 & 83.37 & \textbf{70.20} & \textbf{76.22} & \textbf{95.43} & 81.91 & 88.16 & 85.73 & \textbf{82.69} & \textbf{84.18} \\
\bottomrule
\end{tabular}
\end{table*}

\begin{figure*}[htbp]
  \centering
  \includegraphics[width=\linewidth, page=1]{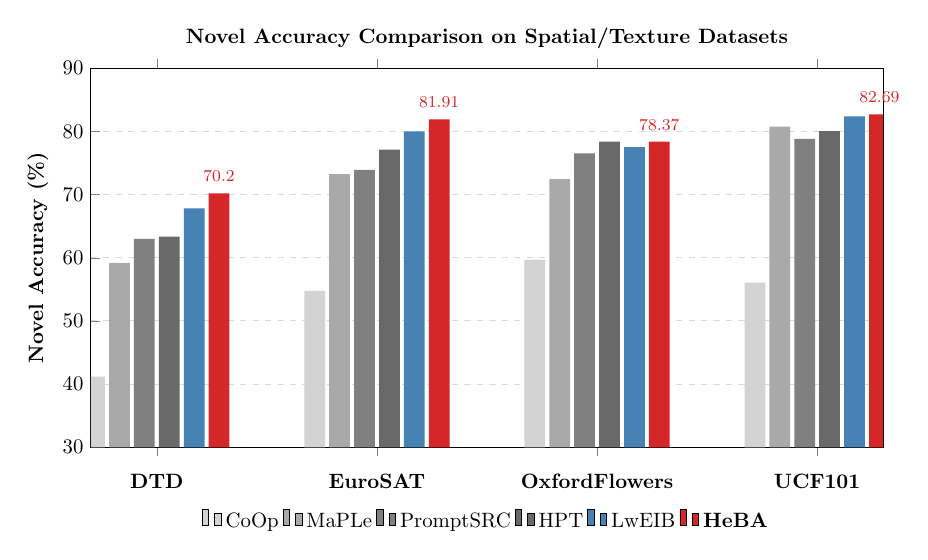}
  \vspace{0.1cm} 
  \includegraphics[width=\linewidth, page=2]{HeBA_barplot.pdf}
  \caption{\doublespacing
  \textbf{Fine-grained Performance Comparison on Structure-Sensitive Datasets.} 
  We report the \textbf{Novel Accuracy} (top) and \textbf{Harmonic Mean} (bottom) on four challenging benchmarks: DTD (textures)~\cite{cimpoi2014describing}, EuroSAT (satellite imagery)~\cite{helber2019eurosat}, Oxford Flowers (fine-grained)~\cite{nilsback2008automated}, and UCF101 (actions)~\cite{soomro2012ucf101}. 
  These domains require capturing local spatial correlations, which standard MLP-based adapters~\cite{gao2023clip} often neglect.
  \textbf{HeBA (Red bars)} consistently outperforms the previous state-of-the-art LwEIB~\cite{yang2025learning} (Blue bars) and other baselines (Gray bars).}
  \label{fig:dataset_comparison}
\end{figure*}

\subsection{Cross-Dataset Evaluation}
\begin{table}[htbp]
\centering
\caption{\doublespacing
Comparison with state-of-the-art methods in the Cross-Dataset Evaluation setting.
Models are trained on ImageNet (16 shots) and evaluated on 10 other datasets.}
\label{tab:cross_dataset}
\footnotesize 
\setlength{\tabcolsep}{3pt} 
\begin{tabular}{lc c cccccccccc c}
\toprule
\multicolumn{1}{c}{Method} & Year & \rotatebox{90}{ImageNet} & \rotatebox{90}{Caltech101} & \rotatebox{90}{OxfordPets} & \rotatebox{90}{StanfordCars} & \rotatebox{90}{Flowers102} & \rotatebox{90}{Food101} & \rotatebox{90}{FGVCAircraft} & \rotatebox{90}{SUN397} & \rotatebox{90}{DTD} & \rotatebox{90}{EuroSAT} & \rotatebox{90}{UCF101} & \rotatebox{90}{\textbf{Average}} \\
\midrule
CLIP~\cite{radford2021learning} & 2021 & 66.72 & 92.98 & 89.13 & 65.29 & 71.30 & 86.11 & 24.90 & 62.59 & 44.56 & 47.84 & 66.83 & 65.15 \\
CoOp~\cite{zhou2022learning} & 2022 & 71.51 & 93.70 & 89.14 & 64.51 & 68.71 & 85.30 & 18.47 & 64.15 & 41.92 & 46.39 & 66.55 & 63.88 \\
CoCoOp~\cite{zhou2022conditional} & 2022 & 71.02 & 94.43 & 90.14 & 65.32 & 71.88 & 86.06 & 22.94 & 67.36 & 45.73 & 45.37 & 68.21 & 65.74 \\
MaPLe~\cite{khattak2023maple} & 2023 & 70.72 & 93.53 & 90.49 & 65.57 & 72.23 & 86.20 & 24.74 & 67.01 & 46.49 & 48.06 & 68.69 & 66.30 \\
P-SRC~\cite{khattak2023self} & 2023 & 71.27 & 93.60 & 90.25 & 65.70 & 70.25 & 86.15 & 23.90 & 67.10 & 46.87 & 45.50 & 68.75 & 65.81 \\
HPT~\cite{wang2024learning} & 2024 & \textbf{71.72} & 94.20 & \textbf{92.63} & 66.33 & \textbf{74.84} & 86.21 & 25.68 & 68.75 & \textbf{50.87} & 47.36 & \textbf{70.50} & 67.74 \\
MMA~\cite{yang2024mma} & 2024 & 71.00 & 93.80 & 90.30 & 66.13 & 72.07 & 86.12 & 25.33 & 68.17 & 46.57 & 49.24 & 68.32 & 66.61 \\
LwEIB~\cite{yang2025learning} & 2025 & 71.31 & 94.51 & 92.50 & \textbf{66.58} & 73.03 & \textbf{86.37} & \textbf{27.70} & \textbf{69.33} & 50.63 & 55.37 & 70.03 & 68.61 \\
\midrule
\textbf{HeBA} & \textbf{Ours} & 71.50 & \textbf{94.81} & 92.20 & 65.41 & 73.04 & 86.13 & 27.09 & 68.22 & 50.71 & \textbf{58.99} & 70.45 & \textbf{68.71} \\
\bottomrule
\end{tabular}
\end{table}
To evaluate the transferability of learned features, we train HeBA on ImageNet \cite{deng2009imagenet} (16 shots) and evaluate it directly on 10 other datasets without fine-tuning. Table \ref{tab:cross_dataset} presents the results.

\textbf{Analysis.} HeBA achieves the highest average accuracy of \textbf{68.71\%} across the 10 target datasets, outperforming LwEIB~\cite{yang2025learning} (68.61\%). Notably, HeBA demonstrates significant robustness in specialized domains. On \textbf{EuroSAT}, HeBA achieves \textbf{58.99\%}, a substantial improvement over LwEIB (55.37\%) and HPT~\cite{wang2024learning} (47.36\%). This \textbf{+3.62\%} gain confirms that our heterogeneous architecture—specifically the spatial adapter with depthwise convolutions—successfully captures domain-agnostic geometric features (e.g., textures, shapes) that transfer well to satellite imagery. We also observe competitive performance on fine-grained tasks like OxfordPets (92.20\%) and Caltech101 (94.81\%), validating that the bottleneck regularizer prevents overfitting to the source domain.

\subsection{Domain Generalization}
We further evaluate the robustness of HeBA on four out-of-distribution (OOD) variants of ImageNet: ImageNet-V2 \cite{recht2019imagenet}, ImageNet-Sketch \cite{wang2019learning}, ImageNet-A \cite{hendrycks2021natural}, and ImageNet-R \cite{hendrycks2021many}. Results are shown in Table \ref{tab:domain_gen}.

\textbf{Analysis.} HeBA maintains strong OOD robustness with an average accuracy of \textbf{60.26\%}, performing comparably to methods like MaPLe (60.27\%) and PromptSRC (60.65\%). Most notably, HeBA achieves the highest performance on \textbf{ImageNet-A} (Adversarial examples) with \textbf{51.36\%}, surpassing MMA (51.12\%), LwEIB (51.00\%), and HPT (50.85\%). This suggests that the \textbf{Active Kaiming Initialization} strategy allows the model to map out robust decision boundaries early in training without collapsing into the source domain's local minima, effectively preserving the backbone's adversarial robustness.

\begin{table}[htbp]
\centering
\caption{\doublespacing
Comparison of domain generalization on ImageNet variants. HeBA shows superior robustness on ImageNet-A (Adversarial).}
\label{tab:domain_gen}
\footnotesize
\setlength{\tabcolsep}{4.5pt} 
\begin{tabular}{lc c cccc c}
\toprule
\multicolumn{1}{c}{Method} & Year & ImageNet & -V2 & -S & -A & -R & \textbf{Average} \\
\midrule
CLIP~\cite{radford2021learning} & 2021 & 66.73 & 60.83 & 46.15 & 47.77 & 73.96 & 57.18 \\
CoOp~\cite{zhou2022learning} & 2022 & 71.51 & 64.20 & 47.99 & 49.71 & 75.21 & 59.28 \\
CoCoOp~\cite{zhou2022conditional} & 2022 & 71.02 & 64.07 & 48.75 & 50.63 & 76.18 & 59.91 \\
MaPLe~\cite{khattak2023maple} & 2023 & 70.72 & 64.07 & 49.15 & 50.90 & 76.98 & 60.27 \\
PromptSRC~\cite{khattak2023self} & 2023 & 71.27 & 64.35 & 49.55 & 50.90 & 77.80 & 60.65 \\
HPT~\cite{wang2024learning} & 2024 & \textbf{71.72} & \textbf{65.25} & 49.36 & 50.85 & 77.38 & 60.71 \\
MMA~\cite{yang2024mma} & 2024 & 71.00 & 64.33 & 49.13 & 51.12 & 77.32 & 60.48 \\
LwEIB~\cite{yang2025learning} & 2025 & 71.31 & 64.47 & \textbf{50.07} & 51.00 & \textbf{77.81} & \textbf{60.84} \\
\midrule
\textbf{HeBA} & \textbf{Ours} & 71.50 & 63.55 & 49.57 & \textbf{51.36} & 76.56 & 60.26 \\
\bottomrule
\end{tabular}
\end{table}
\section{Ablation Study}
To validate the effectiveness of the core components in HeBA, we conduct an ablation study on the Average performance across 11 datasets. The results are summarized in Table \ref{tab:ablation}.

\textbf{Impact of Initialization Strategy.}
We compare our \textbf{Active Kaiming Initialization} strategy against the standard Zero-Initialization used in prior works (`w/ Zero-Init`). While Zero-Initialization achieves a high Novel accuracy (78.63\%), it yields sub-optimal performance on Base classes (84.11\%) due to the delayed convergence typical of zero-gradient initializations. Shifting to an active initialization (HeBA Full) significantly boosts Base accuracy to \textbf{84.29\%} while maintaining comparable Novel performance (78.62\%), resulting in the highest Harmonic Mean of \textbf{81.35\%}. This confirms the theoretical benefit of actively driving feature adaptation from the first iteration.

\textbf{Impact of Spatial Inductive Biases.}
We analyze the contribution of the visual adapter's architecture:
\begin{itemize}
    \item \textbf{w/o Spatial Bias (1D):} Treating image tokens as a flat sequence (removing the 2D reshape) drops the HM to 81.25\%~\cite{dosovitskiy2020image}. This highlights the theoretical imperative of preserving the intrinsic 2D structure of visual data.
    \item \textbf{w/o Depthwise Conv:} Replacing the $3 \times 3$ depthwise convolution with an identity mapping (pointwise only) further degrades performance to 81.20\%~\cite{chollet2017xception}. This structurally validates that local spatial aggregation, provided by the depthwise kernel, is critical for accurately mapping geometric features.
\end{itemize}

\begin{table}[htbp]
\centering
\caption{\doublespacing
Ablation study of HeBA components. \textbf{Full HeBA} (utilizing Kaiming Init and Spatial Depthwise Convolutions) achieves the best trade-off between Base and Novel accuracy.}
\label{tab:ablation}
\footnotesize
\setlength{\tabcolsep}{13pt}
\begin{tabular}{l ccc}
\toprule
\textbf{Configuration} & \textbf{Base} & \textbf{Novel} & \textbf{HM} \\
\midrule
\textbf{HeBA (Full)} & \textbf{84.29} & 78.62 & \textbf{81.35} \\
w/ Zero-Initialization & 84.11 & \textbf{78.63} & 81.28 \\
w/o Spatial Bias (1D) & 84.04 & \textbf{78.63} & 81.25 \\
w/o Depthwise Conv & 83.96 & \textbf{78.63} & 81.20 \\
\bottomrule
\end{tabular}
\end{table}

\subsection{Inference-Time Adapter Scaling}
We further investigate the sensitivity of HeBA to the adapter scaling factor $\alpha$ (denoted as $s$ in Fig.~\ref{fig:heba_structure}) when applied to unseen domains. While the model is trained with a fixed scaling factor $\alpha_{\text{base}} = 0.05$, we hypothesize that the optimal contribution of the adapter varies depending on the severity of the distribution shift. We evaluate varying the inference-time scaling factor $\alpha_{\text{novel}}$.

\textbf{Cross-Dataset Evaluation (Table \ref{tab:ablation_scaling_combined}).}
When transferring to entirely new datasets, \textbf{reducing} the adapter scale improves performance. Setting $\alpha_{\text{novel}} = 0.025$ (half the base scale) yields the highest average accuracy of \textbf{68.71\%}. This suggests that for distinct downstream tasks, dampening the adapter allows the theoretically robust, general-purpose features of the frozen CLIP backbone to take precedence, thereby enhancing transferability.

\textbf{Domain Generalization (Table \ref{tab:ablation_scaling_combined}).}
Conversely, for domain generalization where semantic classes remain consistent (ImageNet variants), the best performance is achieved by maintaining consistency between training and inference scales ($\alpha_{\text{novel}} = \alpha_{\text{base}} = 0.05$), resulting in \textbf{60.67\%}. Reducing the scale hurts performance, indicating that when semantics are shared, the adapter's learned features are highly robust and should be fully utilized.

\begin{table}[htbp]
\centering
\caption{\doublespacing
\textbf{Impact of Inference-time Scaling ($\alpha_{\text{novel}}$).} Models are trained with $\alpha_{\text{base}}=0.05$. Reducing scale helps for Cross-Dataset transfer, while keeping it fixed is optimal for Domain Generalization.}
\label{tab:ablation_scaling_combined}
\footnotesize
\setlength{\tabcolsep}{8pt}
\begin{tabular}{cc cc}
\toprule
\multirow{2}{*}{$\boldsymbol{\alpha_{\text{base}}}$} & \multirow{2}{*}{$\boldsymbol{\alpha_{\text{novel}}}$} & \textbf{Cross-Dataset} & \textbf{Domain Gen.} \\
& & \textbf{Avg Acc (\%)} & \textbf{Avg Acc (\%)} \\
\midrule
0.05 & 0.075 & 68.26 & 60.62 \\
0.05 & 0.050 & 68.60 & \textbf{60.67} \\
0.05 & 0.025 & \textbf{68.71} & 60.26 \\
0.05 & 0.0125 & 68.66 & 59.86 \\
\bottomrule
\end{tabular}
\end{table}
\section{Conclusion}

In this work, we introduced \textbf{HeBA} (Heterogeneous Bottleneck Adapter), a novel parameter-efficient tuning framework for vision-language models that explicitly addresses the modality gap in adaptation. Unlike prior approaches that apply uniform structural priors (e.g., pure MLPs or prompts) across both modalities, HeBA disentangles the adaptation process: it employs a bottleneck linear regularizer to preserve semantic integrity in the text branch and depthwise-separable convolutions to capture local geometric inductive biases in the visual branch.

Extensive experiments on 11 diverse benchmarks demonstrate that HeBA sets a new state-of-the-art for Base-to-Novel generalization, achieving a Harmonic Mean of \textbf{81.35\%}. Crucially, our \textbf{Active Kaiming Initialization} strategy provides a rigorous alternative to standard zero-initialization, demonstrating that ensuring early gradient flow effectively balances plasticity and stability. HeBA exhibits remarkable robustness in cross-dataset transfer and domain generalization settings, particularly on structure-sensitive tasks like satellite imagery (EuroSAT) and textures (DTD), where it outperforms existing methods by significant margins. These results validate that respecting the distinct structural nature of visual and textual modalities is fundamental to robust, generalized few-shot learning.

\section*{Declaration of Interests}
The authors declare that they have no known competing financial interests or personal relationships that could have appeared to influence the work reported in this paper.

\section*{CRediT authorship contribution statement}
\textbf{Md Jahidul Islam:} Conceptualization, Methodology, Software, Validation, Formal analysis, Investigation, Resources, Data curation, Writing - original draft, Writing - review \& editing, Visualization, Project administration.

\section*{Acknowledgements}
The authors declare that no external financial support was received for this research. 
We acknowledge the use of generative AI for language polishing and proofreading 
assistance during the preparation of this manuscript.

\bibliographystyle{elsarticle-num}
\bibliography{reference}
\end{document}